\documentclass[a4paper]{article}

\usepackage{INTERSPEECH2022}
\usepackage{hyperref}

\title{Extending Compositional Attention Networks for Social Reasoning in Videos}
\name{Christina Sartzetaki$^1$, Georgios Paraskevopoulos$^{1,2}$, Alexandros Potamianos$^1$}
\address{
  $^1$School of ECE, National Technical University of Athens, Greece\\
  $^2$Institute for Speech and Language Processing, Athens, Greece}
\email{christina.sartzetaki@gmail.com, \{geopar,potam\}@central.ntua.gr}

\begin{document}

\maketitle
\begin{abstract}
We propose a novel deep architecture for the task of reasoning about social interactions in videos. We leverage the multi-step reasoning capabilities of  Compositional Attention Networks (MAC) \cite{hudson_compositional_2018}, and propose a multimodal extension (MAC-X). 
MAC-X is based on a recurrent cell that performs iterative mid-level fusion of input modalities (visual, auditory, text) over multiple reasoning steps, by use of a temporal attention mechanism.
We then combine MAC-X with LSTMs for temporal input processing in an end-to-end architecture.
Our ablation studies show that the proposed MAC-X architecture can effectively leverage multimodal input cues using mid-level fusion mechanisms.
We apply MAC-X to the task of Social Video Question Answering in the Social IQ dataset and obtain a $2.5\%$ absolute improvement in terms of binary accuracy over the current state-of-the-art.

\end{abstract}

\section{Introduction}
Humans are social creatures; our survival and well-being depends on our effective communication with others. This is achieved through perceiving and understanding information from multiple sensory modalities as well as reasoning and arriving to conclusions, in order to respond accordingly. Artificial intelligence systems need to be able to process interactions between the different sensory modalities to gain an in-depth understanding of their environment, and for that reason multimodal machine learning has developed into a vibrant multi-disciplinary field of increasing importance and extraordinary potential \cite{baltrusaitis_multimodal_2017} with a wide range of benchmark tasks.

In Visual Question Answering (VQA), a task sometimes described as a visual Turing test \cite{manmadhan_visual_2020, xu_fooling_2017}, an AI agent is required to answer a natural language question based on an input image, from answers either in multiple-choice or open-ended format.
The VQA task was introduced in \cite{antol_vqa_2015} and it inspired the creation of several datasets focusing on different aspects of the task \cite{johnson_clevr_2016, goyal_making_2017, agrawal_dont_2018, hudson_gqa_2019}.
The VQA task can also be formulated with video content (Video QA) \cite{jang_tgif-qa_2017, lei_tvqa_2019, grunde-mclaughlin_agqa_2021}, where the input has a temporal dimension and may include audio and dialogue transcript. Video QA is a more complex multimodal task that may require action recognition, conversation and story line understanding, as well as using speech characteristics such as prosody, timbre and pitch.
Social-IQ \cite{zadeh_social-iq_2019} is an unconstrained benchmark that introduces the task of Social Video Question Answering. It consists of human-centered videos in the wild along with social and theory-of-mind-related questions, and answering can demand sophisticated combinations of language understanding, cultural knowledge, logical and causal reasoning, on top of non-social layers of comprehension about physical events \cite{chen_characterizing_2020}.

\begin{figure}[th]
  \centering
  \includegraphics[width=.8\linewidth]{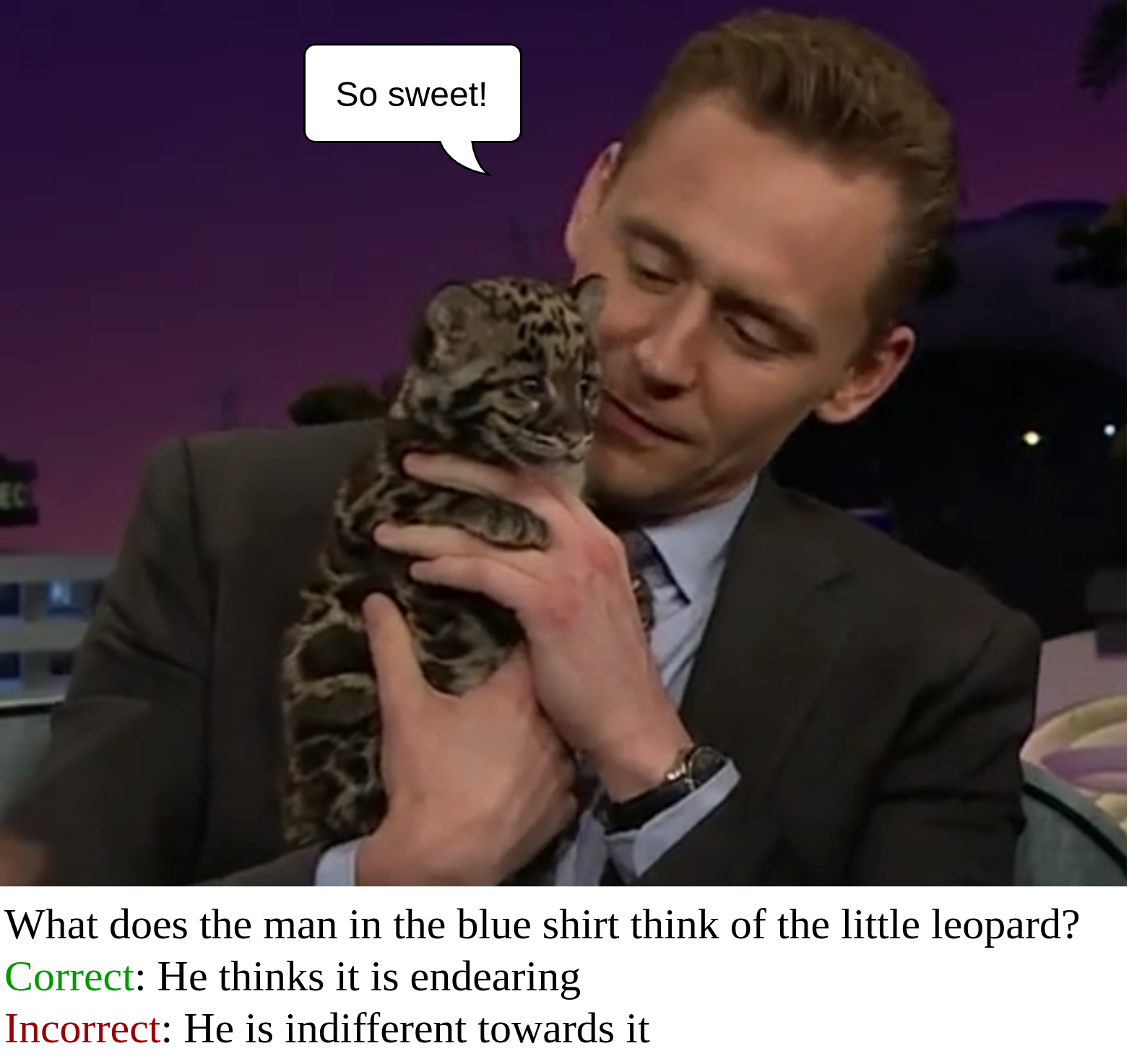}
  \caption{\textbf{Example from the Social-IQ dataset:} The man looks lovingly at the little leopard while exclaiming ``So sweet!"}
\end{figure}


\begin{figure*}[t]
  \centering
  \includegraphics[width=.85\textwidth]{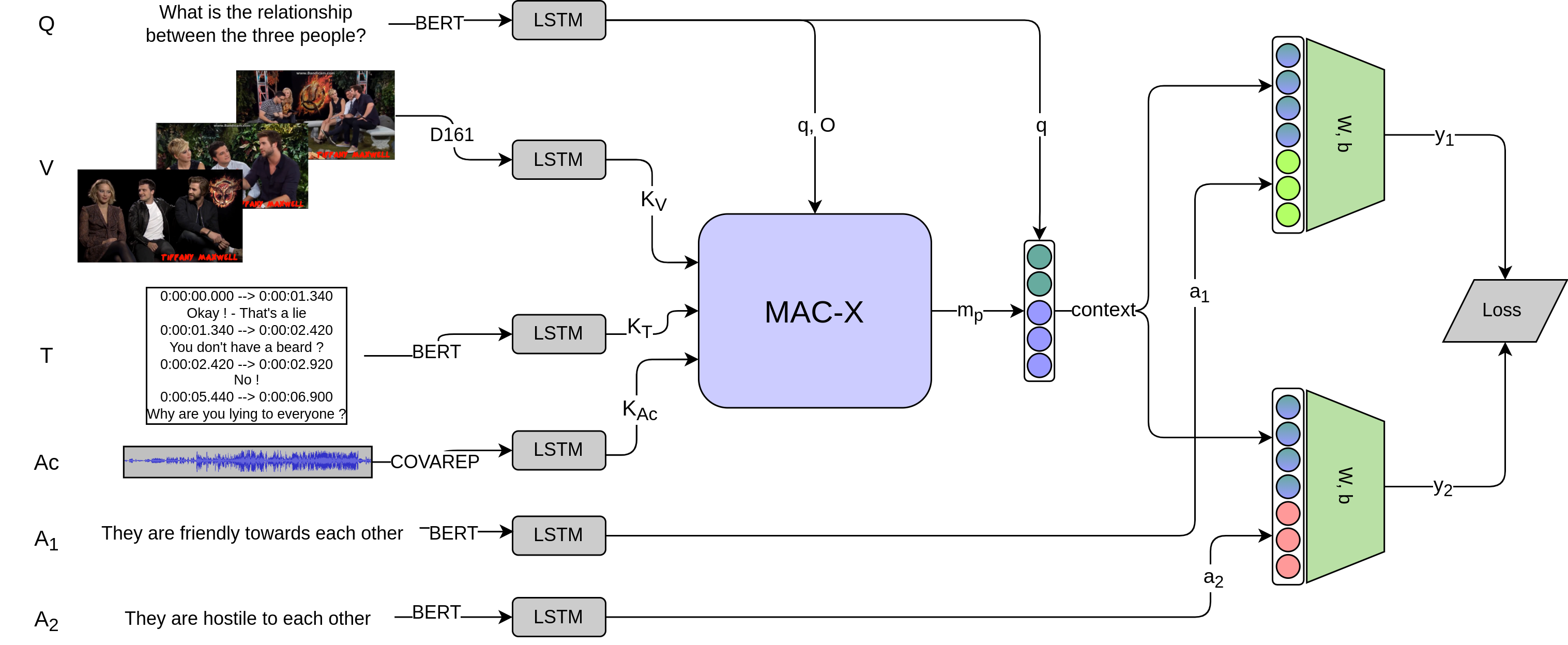}
  \caption{\textbf{Overview of the proposed end-to-end architecture, centered around the MAC-X Network:} On the left, the question ($Q$), visual frames ($V$), dialogue transcript ($T$), acoustic input ($Ac$) as well as correct ($A_1$) and incorrect ($A_2$) answers are shown for the binary task. Their features are encoded with LSTMs, before use in MAC-X or in final classification along with last memory $m_p$. Two identical classifiers make the predictions $y_1, y_2$ which are then used to calculate the loss in equation (\ref{their-loss}).}
  \label{overview}
\end{figure*}

A direction that has proven successful in the VQA literature is combining modules of memory and attention. 
In \cite{xiong_dynamic_2016}, the Dynamic Memory Network (DMN) \cite{kumar_ask_2016} proposed for Text QA is extended for application in VQA, while in \cite{gao_motion-appearance_2018},
it is enhanced with new mechanisms for Video QA.
Notably, \cite{anderson_bottom-up_2018} 
proposes a bottom-up and top-down attention mechanism for salient image regions, and
in \cite{yu_deep_2019}
images and questions are processed through self and cross attention.
Lastly, in \cite{li_beyond_2019}
the commonly used RNNs are replaced with positional self-attention.
Another approach in recent research is neurosymbolic models, which attempt to get the best of both worlds from deep neural networks and older symbolic-AI techniques. 
In \cite{andreas_neural_2016}, 
strong supervision is used to translate questions to functional programs followed by a question-specific neural network, as opposed to \cite{mao_neuro-symbolic_2019}
where this translation requires no explicit supervision.
Moving towards a more neural approach, the method proposed in \cite{hudson_learning_2019}
predicts a probabilistic graph for the image and performs sequential reasoning over the abstract latent space of that graph.

The Memory Attention Composition (MAC) Network \cite{hudson_compositional_2018} was 
proposed in an attempt to capture the ``logic of thought'' in addition to constructing neural representations from the data. The MAC Network exploits the core ideas of attention that underlie neural models, but also provides an architecture suited for soft symbolic reasoning. 
In \cite{le_neural_2020},
the authors introduce a dual process neural architecture for Video QA where MAC is employed as ``System 2'', taking as input a temporal attention space-time representation from ``System 1''.

For the task of Social Video Question Answering, the methods previously explored on Social-IQ typically make use of attention and fusion mechanisms, and can be summarized as follows.
First, Tensor Memory Fusion Network (TMFN) \cite{zadeh_social-iq_2019} is a baseline created by performing architecture and hyperparameter search on TFN \cite{zadeh_tensor_2017} and MFN \cite{zadeh_memory_2018} models and combining them into a joint model, while Multimodal Co-attention based network for Question Answering (MCQA) \cite{kumar_mcqa_2020}
is based on input fusion and alignment, and cross-alignment of joint context with query. 
The RNN-based model in \cite{zhang_temporal_2020} (Temporal Attention and Consistency measuring Network, or TACO-Net)
uses a consistency measurement module in addition to temporal attention, while \cite{gat_removing_2020}
uses a regularization term estimated via the Fisher information to balance the clues between modalities.

In this work, we propose a multimodal extension of MAC Network \cite{hudson_compositional_2018} for Social-IQ, called MAC-Extend (MAC-X). 
The motivating factors for this approach are that MAC: 1) was intended for tasks that require deliberate reasoning from facts to conclusions \cite{hudson_compositional_2018} on account of its structured and iterative reasoning, and 2) consists of thoroughly general-purpose modules and operations.
We believe that these characteristics make it very well-suited for Social-IQ, and a strong baseline for the task of Social Reasoning as well as any reasoning task.\\

\textbf{Our main contributions are:}
\begin{itemize}
    \item We present MAC-X, a multimodal extension of the MAC Network, featuring temporal attention, a mid-level fusion mechanism, and multiple-choice Video Question Answering capabilities.
    \item We analyse the performance of MAC-X in the challenging Social-IQ dataset through ablation studies and comparison to prior state-of-the-art methods, which it significantly outperforms.
    \item Our code is publicly available \footnote{https://www.github.com/SergeantChris/macx-socialiq}.
\end{itemize}

\section{Our Approach: MAC-X}

\subsection{Overview}
Our model is based on the MAC Network, a recurrent architecture of length $p$ and dimension $d$ defined by the Memory, Attention and Composition (MAC) cell which performs an attention-based reasoning step $i$ given a knowledge base and a query. The MAC cell is composed of three operational units, the Control Unit, the Read Unit, and the Write Unit. This pipeline reads from input features in a way that is controlled by part of the query and memory from previous readings, proceeding to incorporate that into the current memory. 

Building on these 
structural priors, MAC-X extracts information from multiple sources, formulates its attention over time instead of space, performs a mid-level fusion on the intermediate representations of the modalities, and ultimately facilitates multiple-choice Question Answering on multimodal data. An overview of the model's architecture for the task of Social Video QA can be seen in Figure \ref{overview}, and the enhanced cell's architecture is shown in Figure \ref{cell}.
In the following sections, all equations and figures are described for the binary task for simplicity, and can be directly extended for the multiple choice task in which we also report results. More details on the two tasks in Section \ref{setup}.

\subsection{Input Units}
As shown in Figure \ref{overview}, the language modality inputs which consist of the question ($Q$), the dialogue transcript ($T$) and the correct and incorrect answers ($A_1$, $A_2$ respectively), are initially encoded with last hidden state BERT embeddings, while the visual modality ($V$) with Densenet161 (D161) features for each frame (at 1fps), and the acoustic modality ($Ac$) with COVAREP features. They are then passed through bidirectional LSTMs whose outputs constitute the knowledge bases \(K_V\), \(K_T\) and \(K_{Ac}\) for the visual, transcript and acoustic input respectively and the contextual words $O$ for the question. The last hidden states \(q\), \(a_1\), and \(a_2\) are used as the vector representation for the question and answers respectively. The output dimension of the LSTMs is $d$, where $d$ is the dimension of the MAC model. Each of the knowledge bases can be described as \(K_j^{L\times d} = \{k_t|_{t=1}^{L}\}\), where $L$ is the sequence length of modality $j$ in the time dimension $t$.

\subsection{Control Unit}
\begin{figure}[th]
  \centering
  \includegraphics[width=\linewidth]{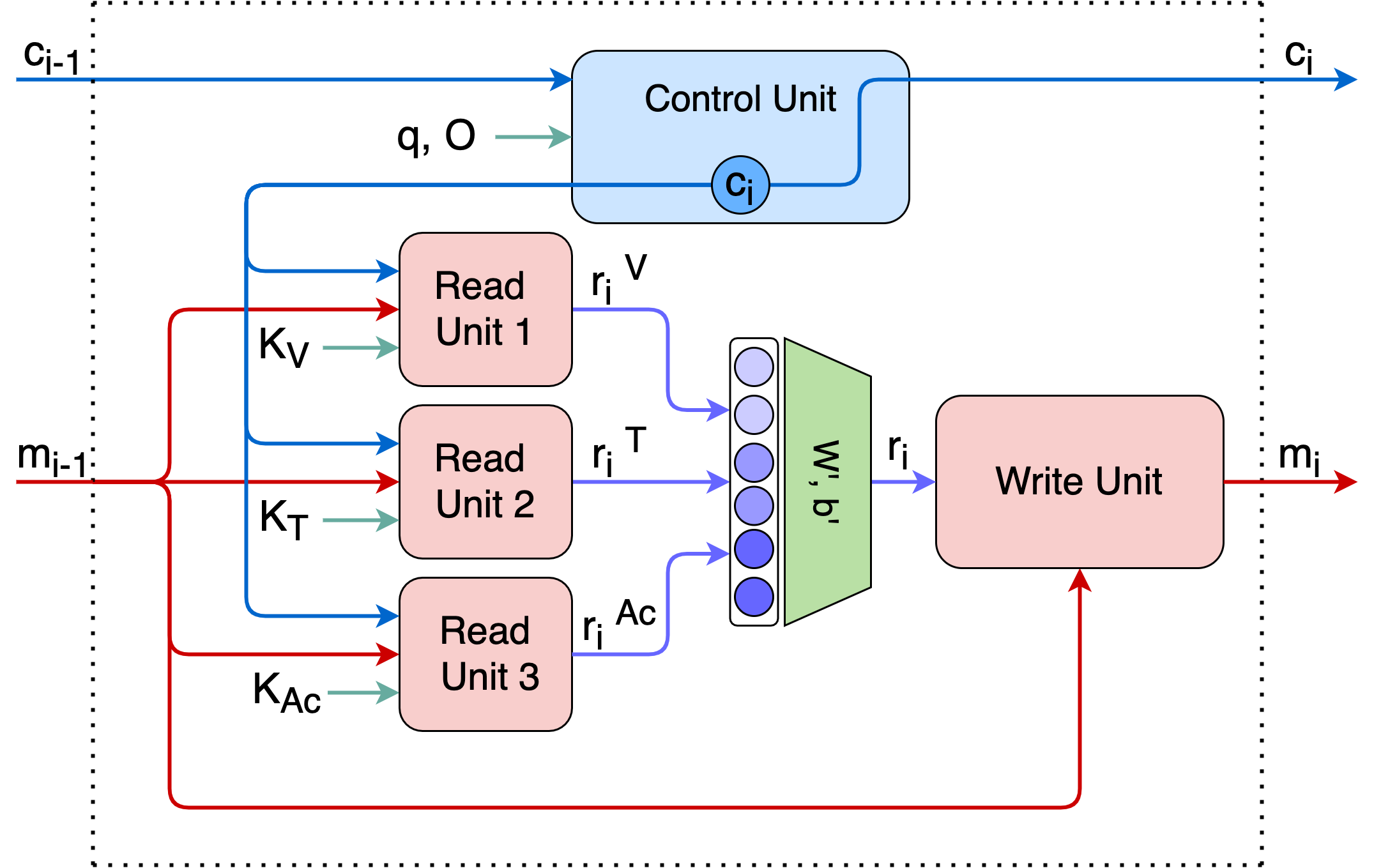}
  \caption{\textbf{The MAC-X recurrent cell in the $i$\,th reasoning step:} The multimodal extension of the MAC cell is manifested in the cloning of the Read Unit and consequent fusion of the modalities' extracted information $r_i^j$ before integration to memory $m_i$.}
  \label{cell}
\end{figure}

The Control Unit (Figure \ref{cell}) stays the same as in the original architecture, and can be summarized as
\begin{equation}
    c_i = \sum_{s=1}^S\sigma(f_c(f_{cq}([c_{i-1}, f_q(q)])\odot O_s))\cdot O_s
\end{equation}
where S is the number of contextual words, $\sigma$ the softmax function, and $f_x$ 
are single layer feedforward networks. In the equation above, attention is performed on the contextual words $O$ based on information from the question $q$ and the previous control $c_{i-1}$, in order to update the current $c_i$. This $c_i$ determines what part of the question we want to extract knowledge about from the input modalities in the current reasoning step.

\subsection{Multiple Read Units}
For reading from the knowledge bases, a simple cloning of the Read Unit for each modality is proposed, each getting a copy of the current control and previous memory (see Figure \ref{cell}).
This approach allows for the control $c_i$ to attend to the different modalities independently at the same reasoning step, while at the same time being conditioned on a memory that is kept collectively for all of them. For example, previous information from the audio and visual modalities could be important to determine the next most useful information to integrate from the transcript. The operation of each Read Unit $j$ is defined as
\begin{equation}
    I_{i,t}^j = f_{mk}([f_m(m_{i-1})\odot f_k(k_t^j), k_t^j])
    \label{information}
\end{equation}
\begin{equation}
    r_i^j = \sum_{t=1}^L\sigma(f_r(c_i\odot I_{i,t}^j))\cdot k_t^j
    \label{read}
\end{equation}
where $j = {V, T, Ac}$ are the different modalities. In the former of the above equations, information $I_{i, t}^j$ is gathered from the knowledge base of modality $j$ at each position $t$ in its temporal sequence. This information is considered to be only optionally related to the previous memory $m_{i-1}$, and so the initial $k_t^j$ is also concatenated in the input vector of equation (\ref{information}). In equation (\ref{read}), attention based on the current control $c_i$ is performed on $k_t^j$, to create the current $r_i^j$ for each Read Unit.

\subsection{Multimodal Fusion}
In order to perform a mid-level fusion, we fuse modalities at this stage by concatenating the intermediate extracted knowledge results $r_i^j$ for every modality $j$ and passing them through a feedforward layer, effectively constructing a single shared representation layer $r_i$ for all modalities. This is shown in Figure \ref{cell} and in the equation
\begin{equation}
    r_i = W'[r_i^V, r_i^T, r_i^{Ac}] + b'
\end{equation}
Implementing the multimodal fusion at this innermost stage stands in contrast to simpler late fusion methods, a comparison discussed in detail in Section \ref{results}.

\subsection{Write Unit} \label{write_unit}
The Write Unit (Figure \ref{cell}) integrates the collective information $r_i$ from the Read Units to the previous memory $m_{i-1}$ 
and thus obtains the current memory $m_i$.
\begin{equation}
    m_i = f_{mr}([m_{i-1}, r_i])
\end{equation}
In this work we omit the optional components of the Write Unit proposed in \cite{hudson_compositional_2018}, as the authors' experiments suggest that their use does not yield significant improvements.

\subsection{Output Unit}
After $p$ recurrent iterations of the MAC-X cell as described in the previous sections, the final memory $m_p$ is concatenated with the question representation $q$ to create the context on which the correct answer should be chosen (Figure~\ref{overview}). This is further concatenated with each of the answers $a_1, a_2$ and passed to identical two layer feedforward networks for classification, which output the predictions
\begin{equation}
    y_1 = W[q, m_p, a_1]+b, \quad y_2 = W[q, m_p, a_2]+b 
\end{equation}
where $y_1$ and $y_2$ are the correct and incorrect answer predictions respectively. 
We then compute the composite loss
\begin{equation}
    \mathcal{L} = (\frac{1}{N}\sum_{i=1}^{N}y_1^i-1)^2 + (\frac{1}{N}\sum_{i=1}^{N}y_2^i)^2
    \label{their-loss}
\end{equation}
where $N$ is the number of samples in a batch.
We note that this is the same loss that is exhibited in the original code released for the Social-IQ baseline in \cite{zadeh_social-iq_2019}.
The binary accuracy A2 is formulated as
\begin{equation}
    A2 = \frac{1}{M}\sum_{i=1}^{M}(y_1^i>y_2^i)
    \label{acc_strict}
\end{equation}
where M is the total number of samples in the set for which the accuracy is calculated.

\section{Experimental Setup} \label{setup}
\noindent\textbf{Dataset:} The Social IQ dataset (public release) contains 1015 videos, with six questions corresponding to each video and each question having four correct and three incorrect candidate answers. The training set contains 888 videos and the validation set $127$ ($87\%$ - $13\%$ split). 
In all experiments the above validation set is used for evaluation and comparison of the models, as the private test set is reserved
by the authors for future challenges. 
For all input modalities, we use the precomputed embeddings published in \cite{zadeh_social-iq_2019}.

\noindent\textbf{Hyperparameters:} For the LSTM baseline, after all modalities are encoded, they are concatenated and passed directly to the classifiers for final prediction. All experiments with the TMFN baseline are reproduced on the validation set, and the original code released is used. For our model (MAC-X), hyperparameters are set as $p=12$, $d=512$, and no optional self-attention or memory gate mechanisms from \cite{hudson_compositional_2018} are used, as mentioned in Section \ref{write_unit}. All LSTMs are bidirectional, with output dimension $d$ for use in the MAC-X cell. For the comparison to previous state-of-the-art models in Table \ref{sota}, we use their reported results on the validation set.
In all experiments, models are trained on 32 samples per batch, with Adam optimizer and learning rate of $10^{-3}$, for $10$ epochs for LSTM and MAC-X and $50$ epochs for TMFN.
Our models are trained for approximately 5 hours on one NVIDIA GeForce GTX 1080 Ti GPU.

\noindent\textbf{Evaluation:} The dataset metrics are binary (A2) and four-way (A4) accuracy for the binary and multiple choice tasks respectively, following the original formulation presented in \cite{zadeh_social-iq_2019}. For the binary task (A2) we take all $12$ combinations of correct and incorrect answers for a question, resulting in a dataset of $73,080$ total samples where the goal is to select the correct answer between the two. For the multiple choice task (A4) we take all four combinations of one correct and three incorrect answers for a question, resulting in a total of $24,360$ samples where the goal is to select the single correct answer from four choices. Note, the performance of random choice is $50\%$ for A2 and $25\%$ for A4.

\section{Results and Discussion} \label{results}
We next show the results for the proposed architecture and reproduced baselines. All results are averaged over five runs. Input modalities are denoted as $Q$ for the question, $A$ for the answers, $V$ for the visual frames, $T$ for the dialogue transcript, and $Ac$ for the acoustic input.

In Table \ref{mods_ablation} we compare our model (MAC-X) to the LSTM and Tensor Memory Fusion Network (TMFN) \cite{zadeh_social-iq_2019} baselines based on the binary accuracy (A2), in an ablation study for different combinations of the input modalities; each combination is denoted by the modalities it makes use of.
It is observed that in both baselines multimodality is not necessarily beneficial to performance, and can even degrade it substantially. In contrast, MAC-X performs best when all modalities are used, marking a $0.25\%$ absolute accuracy improvement over its single modality input counterparts, which points to the soundness of its knowledge extraction and fusion methods. At the same time it is very effective in the unimodal input settings, surpassing both the LSTM and TMFN baselines by at least five percentage points.
As for the observed importance of each modality, the visual and audio modalities seem to perform best in the LSTM and TMFN baselines respectively, while MAC-X benefits fairly equally from all modalities.
In addition, we show that using just the question and answer (or even just the answer) modalities in the LSTM baseline achieves performance well above random, attesting to the existence of language bias in the validation set.

\begin{table}[th]
  \caption{Ablation study on input modalities and comparison to baseline models, reporting on A2 validation set accuracy}
  \label{mods_ablation}
  \centering
  \begin{tabular}{ l c c c}
    \toprule
    \multicolumn{1}{c}{Mod/ties} & 
                        \multicolumn{1}{c}{\textbf{LSTM}} & \multicolumn{1}{c}{\textbf{TMFN}} & \multicolumn{1}{c}{\textbf{MAC-X}}  \\
    \midrule
    A                       & 63.22 (\textpm 0.41) &   -  & - \\
    QA                      & 64.51 (\textpm 0.58) &   -  & - \\
    QAV                     & 64.82 (\textpm 0.67) & 65.67 (\textpm 0.38) & \textbf{71.01} (\textpm 0.24) \\
    QAT                     & 64.54 (\textpm 0.57) & 65.51 (\textpm 0.43) & \textbf{70.97} (\textpm 0.44) \\
    QAAc                    & 64.17 (\textpm 0.32) & 65.89 (\textpm 0.32) & \textbf{71.00} (\textpm 0.30) \\
    \midrule
    QAVTAc\!   & 63.73 (\textpm 0.71)  & 65.62 (\textpm 0.55) & \textbf{71.25} (\textpm 0.15)   \\
    \bottomrule
  \end{tabular}
\end{table}

In Table \ref{stage} we present an ablation study that showcases the effectiveness of our mid-level fusion method, outperforming a late fusion baseline in both metrics. In the latter's setting, each modality goes through a completely separate MAC Network, whose outputs are fused at that late stage in the same manner as in our mid-level fusion, before entering the final classifiers. This indicates the advantage of fusing modalities at the intermediate representation stage in the models, where their collective useful information can be jointly processed further.

\begin{table}[th]
  \caption{Ablation study on the multimodal fusion stage, reporting on the validation set with the full set of input modalities}
  \label{stage}
  \centering
  \begin{tabular}{ l  c  c}
    \toprule
    \multicolumn{1}{c}{Models} & 
                        \multicolumn{1}{c}{\textbf{A2}} &
                        \multicolumn{1}{c}{\textbf{A4}} \\
    \midrule
    MAC w. Late fusion & 70.59 (\textpm 0.62) & 46.46 (\textpm 0.26) \\
    MAC-X & \textbf{71.25} (\textpm 0.15) & \textbf{47.22} (\textpm 0.60) \\
    \bottomrule
  \end{tabular}
\end{table}
In Table \ref{sota} we measure the performance of our proposed model against five prior state-of-the-art methods, reporting on both metrics for the validation set. We observe a $2.3-2.6\%$ accuracy improvement from the previous state-of-the-art in the binary task (MCQA \cite{kumar_mcqa_2020}), taking variance into account. 
As regards the multiple choice task (A4), we obtain comparable results to the best-performing model TACO-Net \cite{zhang_temporal_2020}. Note that TACO-Net
measures explicitly the consistency between each answer and modality, contributing to the robustness of the model in the multiple choice setting. 
Overall, through implementing and applying MAC-X we set a new leading performance for the binary task of the Social-IQ dataset.

\begin{table}[th]
  \caption{Performance comparison to state-of-the-art methods on the Social-IQ validation set. We report averaged results and standard deviation over five runs.}
  \label{sota}
  \centering
  \begin{tabular}{ l  c  c}
    \toprule
    \multicolumn{1}{c}{Models} & 
                        \multicolumn{1}{c}{\textbf{A2}} &
                        \multicolumn{1}{c}{\textbf{A4}} \\
    \midrule
    TMFN \cite{zadeh_social-iq_2019} & 65.62 & 36.24 \\
    Removing bias \cite{gat_removing_2020} & 67.93 & - \\
    TACO-Net \cite{zhang_temporal_2020} & 68.19 & \textbf{49.08} \\
    Perceptual score \cite{gat_perceptual_2021} & 68.65 & - \\
    MCQA \cite{kumar_mcqa_2020} & 68.80 & 38.30 \\
    \midrule
    Ours (MAC-X) & \textbf{71.25} (\textpm 0.15) & 47.22 (\textpm 0.60) \\
    \bottomrule
  \end{tabular}
\end{table}

\section{Conclusions}
We present MAC-X, a multimodal extension of the MAC Network capable of handling complex multiple choice and multiple modality reasoning tasks like Social-IQ, where we evaluate it and obtain state-of-the-art results.
We conclude that structural priors as well as compositional reasoning can prove useful to Social Video Question Answering, in which - to the best of our knowledge - this direction is applied for the first time. 
We can further confirm from our ablation studies that MAC-X can effectively benefit from all modalities and that mid-level fusion performs considerably better than the late fusion baselines.

A limitation of our system is that it depends only on the precomputed features provided for the visual, audio, and text modalities, and therefore excludes the incorporation of additional sources such as explicit emotion recognition, object-level features, and external knowledge, which correspond to traits that regularly appear in Social-IQ.
In the future, we plan to investigate enhancing the input with the use of such auxiliary features, experiment with more sophisticated techniques of mid-level fusion for the purpose of learning better intermediate multimodal representations, as well as explore a more tailored modelling of the multiple choice task.

\bibliographystyle{IEEEtran}
\bibliography{myfinalbib}

\end{document}